\pdfoutput=1

\documentclass{article}
\usepackage{ijcai22}

\usepackage{times}

\usepackage{soul}
\usepackage{url}
\usepackage[hidelinks]{hyperref}
\usepackage[utf8]{inputenc}
\usepackage[small]{caption}
\usepackage{graphicx}
\usepackage{amsmath}
\usepackage{booktabs}
\newcommand{\squishbegin}{
 \begin{list}{$\bullet$}
  { \setlength{\itemsep}{0pt}
     \setlength{\parsep}{1pt}
     \setlength{\topsep}{1pt}
     \setlength{\partopsep}{0pt}
     \setlength{\leftmargin}{1.5em}
     \setlength{\labelwidth}{1em}
     \setlength{\labelsep}{0.5em} 
  } 
}

\newcommand{\squishtwobegin}{
 \begin{list}{$-$}
  { \setlength{\itemsep}{1pt}
     \setlength{\parsep}{1pt}
     \setlength{\topsep}{1pt}
     \setlength{\partopsep}{0pt}
     \setlength{\leftmargin}{1.5em}
     \setlength{\labelwidth}{1em}
     \setlength{\labelsep}{0.5em} 
  } 
}

\newcommand{\squishend}{
  \end{list}  
}

\newcommand{\btable}[1]{\begin{table}[#1] \begin{center} }
\newcommand{\etable}[2]{\end{center} \vspace{-5pt} \caption{#2} \label{#1} \vspace{-15pt}\end{table}}

\newcommand{\wbtable}[1]{\begin{table*}[#1] \begin{center} }
\newcommand{\wetable}[2]{\end{center} \caption{#2} \label{#1} \end{table*}}

\newcommand{\bfigure}[1]{\begin{figure}[#1]}
\newcommand{\efigure}[2]{\vspace{-8pt} \caption{#2} \label{#1} \end{figure}}

\usepackage{amssymb}
\usepackage{amsthm}
\usepackage{mathtools}
\usepackage{comment}
\usepackage{amsmath}
\usepackage{amsfonts}
\newcommand{\Kate}[1]{\textcolor{red}{Kate: #1}}
\usepackage[table]{xcolor}

\makeatletter
\newtheorem*{rep@theorem}{\rep@title}
\newcommand{\newreptheorem}[2]{%
\newenvironment{rep#1}[1]{%
 \def\rep@title{#2 \ref{##1}}%
 \begin{rep@theorem}}%
 {\end{rep@theorem}}}
\makeatother

\newtheorem{prop}{Proposition}
\newreptheorem{prop}{Proposition}
\newreptheorem{theorem}{Theorem}

\newtheorem{definition}{Definition}

\newcommand{\inteam}{\nu}

\title{Exploring the Benefits of Teams in Multiagent Learning\\
\vspace{-60pt}
\large International Joint Conference on Artificial Intelligence (IJCAI-ECAI) 2022 
\vspace{50pt}
}

\author{
David Radke$^1$\footnote{Contact Author}\and
Kate Larson$^1$\And
Tim Brecht$^{1}$\\
\affiliations
$^1$David R. Cheriton School of Computer Science, University of Waterloo\\
\emails
\{dtradke, kate.larson, brecht\}@uwaterloo.ca
}

\begin{document}

\maketitle

\begin{abstract}

For problems requiring cooperation, many multiagent systems implement solutions among either individual agents or across an entire population towards a common goal.
Multiagent teams are primarily studied when in conflict; however, organizational psychology (OP) highlights the benefits of teams among human populations for learning how to coordinate and cooperate.
In this paper, we propose a new model of multiagent teams for reinforcement learning (RL) agents inspired by OP and early work on teams in artificial intelligence.
We validate our model using complex social dilemmas that are popular in recent multiagent RL and find that agents divided into teams develop cooperative pro-social policies despite incentives to not cooperate.
Furthermore, agents are better able to coordinate and learn emergent roles within their teams and achieve higher rewards compared to when the interests of all agents are aligned.

\end{abstract}

\section{Introduction}

Observed in both animal and human behavior, the ability to work in teams can magnify a group's abilities beyond the capability of individuals.
Organizational psychology (OP) and biology have done extensive research on how ``teams-of-teams" with a shared overall goal increase the collective efforts of all component teams and individuals~\cite{Zaccaro2020MultiteamSA}.
Recently, there is increasing interest in making the study of cooperation central to the development of artificial intelligence (AI) and multiagent systems (MAS)~\cite{DafoeNature2021}.
We propose that adapting a similar team structure as in OP
to a population of AI agents serves as a middle ground between centralized and decentralized systems of coordination and cooperation that can benefit agents' learning processes.



In the context of teams, multiagent reinforcement learning (MARL) has achieved impressive results in competitive two-team zero-sum settings such as capture the flag~\cite{Jaderberg2019HumanlevelPI}, 
hide-and-seek~\cite{Baker2020EmergentTU}, and Robot Soccer (RoboCup)~\cite{Kitano1997RoboCupTR}.
However, when agents are deployed into the real world, they will be faced with problems that are not zero-sum~\cite{Baker2020EmergentRA}.
Therefore, it is essential to explore cooperation in mixed-motive domains, such as Sequential Social Dilemmas (SSDs)~\cite{Leibo2017MultiagentRL}.
With the growing interest in exploring mixed-motive domains
and investigating the impact of group size and structure on system stability~\cite{Nisioti2020GroundingAI}, we aim to model teams and understand their benefits on MARL agents' ability to learn.

Inspired by group structures in OP and early models of teams from the AI literature for task completion, we propose a general model of multiagent teams and validate it in the context of social dilemmas.
It is well documented that individual RL agents fail to learn cooperation in social dilemmas while agents with common interest have more success~\cite{Anastassacos2020PartnerSF,Baker2020EmergentTU}.
Our teams model is situated between these two extremes, where teammates are bound by common interest but mixed-motives exist between non-teammates.
We show, across several games, that teams improve how agents learn and develop pro-social policies.
This work makes the following contributions:

\begin{itemize}
    \item We define a model of teams inspired by early work in multiagent systems and organizational psychology.
    
    \item We discuss the theoretical ramifications of our model in the context of social dilemmas regarding game theoretical incentives under different environmental conditions.

    \item Through an extensive empirical evaluation, we show how our model of teams helps agents develop globally beneficial pro-social behavior despite short-term incentives to defect. As a result, agents in these teams achieve higher rewards in complex domains than when the interests of all agents are aligned, and autonomously learn more efficient combinations of roles when compared with common interest scenarios.
\end{itemize}

\section{Related Work}
\label{sec:related_work}




There is a long history of exploring how agents can coordinate their behavior via teamwork, typically for task completion domains.
Early work by Pollack formally defines an agent's mental model for making collaborative plans between two agents
~\cite{Pollack90plansas}.
Extending Pollack's work, \cite{Grosz1988PlansFD} construct SharedPlans, a model that includes more actions and agent beliefs, similar to shared mental models in human teamwork and modeled joint actions as non-additive~\cite{Grosz1996CollaborativePF}.
Tambe uses a task allocation structure similar to SharedPlans in developing STEAM, a general model of teams where tasks can be completed by sub-teams of agents within a larger system~\cite{Tambe1997TowardsFT}.
The idea of sub-teams within a population was novel to AI;
however, SharedPlans and STEAM remained unable to broadly generalize due to limited agent capabilities.

Recent work with teams in MAS has focused on ad hoc teamwork~\cite{Macke2021ExpectedVO}, 
teams in competition~\cite{Ryu2021CooperativeAC}, 
or coordination problems~\cite{Jaques2019SocialIA}.
When not in competition, teams in AI are typically designed to achieve a common objective and has been criticized for its disregard for adapting significant findings in OP~\cite{Andrejczyk2017Concise}.
We evaluate our model of teams in other settings where teams are not directly competitive, complex mixed-motive social dilemmas, to explore how structures inspired by OP can be adapted to MAS.

Studying social dilemmas and how agents or people can overcome them has been a topic of research in game theory, economics, psychology, and more recently, AI.
Fostering cooperation in social dilemmas with MARL agents often relies on reward sharing among the population~\cite{McKee2020SocialDA} or only a subset of agents with uncertainty~\cite{Baker2020EmergentRA}.
We broadly classify the literature into centralized and decentralized approaches of fostering cooperation.

Centralized systems have taken various forms in the literature, ranging from centralized training for better coordination at test time~\cite{Kraemer2016MultiagentRL}
to explicitly providing access to another agent's internal state~\cite{Deka2020NaturalEO}.
While centralized systems are efficient and convergence is reliable, it is often unsafe to assume access to all agents and they are brittle to exogenous changes.

Inspired by the hypothesized emergence of cooperation in humans and evolutionary game theory, decentralized systems that promote cooperation are primarily implemented at the individual agent level.
Giving agents the ability to punish and sanction others in response to a specific behavior has been shown to foster cooperation in MARL~\cite{Anastassacos2021CooperationAR,Leibo2017MultiagentRL}.

In this work, we adapt findings from OP to AI by constructing a general model of multiagent teams inspired by human teams, SharedPlans, and STEAM to analyze the benefits of teams on how MARL agents learn.
We divide a population of MARL agents into teams not in direct competition and show how teams, as they have been found to do with humans, improve how RL agents co-evolve and learn in challenging domains.
We position our work between centralized and decentralized systems. 
While our teams model does not assume any form of centralized control, the team structure itself provides a way for agents to better coordinate.

\section{A Model for Multiagent Teams}

We model our base environment as a stochastic game $\mathcal{G}=\langle N, S, \{A\}_{i\in N}, \{R\}_{i\in N}, P, \gamma, \Sigma \rangle$.
$N$ is our set of all agents that learn online from experience and $S$ is the state space, observable by all agents, where $s_i$ is a single state observed by agent $i$.
$A = A_1\times \ldots \times A_N$ is the joint action space for all agents where $A_{i}$ is the action space of agent $i$.
$R = R_1 \times \ldots \times R_N$ is the joint reward space for all agents where $R_{i}$ is the reward function of agent $i$ defined as $R_i: S \times A\times S \mapsto \mathbb{R}$, a real-numbered reward for taking an action in an initial state and resulting in the next state.
$P:S\times A\mapsto \Delta(S)$ represents the transition function which maps a state and  joint action into a next state with some probability and $\gamma$ represents the discount factor so that $0 \leq \gamma < 1$.
$\Sigma$ represents the policy space of all agents, and the policy of agent $i$ is represented as $\pi_i:S \mapsto A_{i}$ which specifies an action that the agent should take in an observed state.\footnote{We can also allow for randomized policies.}

Our teams model consists of a stochastic game with teams
 $\langle \mathcal{G}, \mathcal{T} \rangle$, where
$\mathcal{T}$ is a  partition of the population of agents into disjoint teams,  $\mathcal{T} = \{ T_i | T_i \subseteq N, \cup T = N, T_i \cap T_j = \emptyset \forall i, j \}$.
A team \emph{structure} defines the composition of $\mathcal{T}$, such as the number of teams and agents on each team.
Consistent with the original groundwork on multiagent teams~\cite{Tambe1997TowardsFT,Grosz1988PlansFD}, we define a team of agents as being bounded together through common interest.
Consistent with recent MARL work, we model common interest through reward sharing making the assumption agents value rewards equivalently~\cite{McKee2020SocialDA,Hughes2018InequityAI}.
We define a new reward function for agents on a team as $TR_i:S\times A\times S \mapsto \mathbb{R}$ so that the reward for $i \in T_i$ is dependant on their own behavior and that of their teammates.
Any function can be implemented to define $TR_i$.
In our analysis and experiments, we use:

\begin{equation}
    TR_i=\frac{\sum_{j\in T_i} R_j(S,A,S')}{|T_i|},
    \label{eq:team_reward}
\end{equation}

\noindent
where teammates share their rewards equally to be consistent with past work~\cite{Wang2019EvolvingIM,Baker2020EmergentTU}.

Agents learn from their individual experience using RL.
As is standard in many MARL problems, agents are trained to independently maximize their own rewards. 
In particular, at time $t$ each agent $i$ selects some action $a_i$ which together form a joint action $a^t$.  
This action results in a transition from state $s^t$ to state $s^{t+1}$, according to the transition function $P$, and provides each agent $i$ with reward $R_{i,t}(s^t,a^t,s^{t+1})$. 
Agents seek to maximize their sum of discounted future rewards, $V_i=\sum_{t=0}^\infty \gamma^t R_{i,t}$.
Our model replaces $R_{i}$ with $TR_{i}$, reconfiguring the learning problem so that agents must simultaneously learn what individual behavior maximizes their team's expected discounted future reward.

\section{Social Dilemmas}


Social dilemmas are situations which present tension between short-term individual incentives and the long-term collective interest where agents must cooperate for higher rewards.
All agents prefer the benefits of a cooperative equilibria; however, the short-term benefits of self-interested behavior outweigh those of cooperative behavior.
For our analysis, we consider intertemporal social dilemmas with active provision defined as when cooperation carries an explicit cost~\cite{Hughes2018InequityAI}.
We implement our model of teams in the Iterated Prisoner's Dilemma (IPD)~\cite{Rapoport1974PrisonersD} and the Cleanup domain game~\cite{SSDOpenSource}.

\subsection{Environment 1: Iterated Prisoner's Dilemma}

In the one-shot Prisoner's Dilemma, two agents interact and each must decide to either cooperate ($C$) with or defect ($D$) on each other.
We assume there is a cost ($c$) and a benefit ($b$) to cooperating where $b>c>0$. 
If an agent cooperates, it incurs the cost $c$. 
If both agents cooperate, they both also benefit, each receiving a reward of $b-c$.
If one agent cooperates but the other defects, then we assume that the cooperating agent incurs the cost $c$, but the defecting agent reaps the benefit $b$ (e.g., by stealing the contribution of the cooperator).
If neither cooperate, neither benefit nor incur a cost, leading to a reward of zero for both.
The unique Nash Equilibrium is obtained when both agents defect, represented by  ($D$, $D$).
Joint cooperation does not form an equilibrium, since if one agent cooperates, the other agent is strictly better off defecting and receiving $b$, instead of $b-c$.


In the IPD, this game is repeatedly played which adds a temporal component and allows agents to learn a policy over time.
Instead of just two agents, we work with a population of agents that are divided into teams \emph{a priori}.
At each timestep, agents are randomly paired with another agent, a \emph{counterpart}, that may or may not be a teammate.
Agents are informed as to what team their counterpart belongs to through a numerical signal $s_i$, though additional identity information is not shared.
Agents must decide to cooperate with or defect on their counterpart. 
Their payoff for the interaction is the team reward, $TR_i$, based on their own and other teammates' interactions.
Agents update their strategies (i.e., learn) using their direct observation $s_i$, what action they chose $a_i$, and their team reward $TR_i$. 
Since only the team information of the counterpart is shared, the strategies of all agents on team $T_i$ ultimately affects how agents learn to play any member of $T_i$.

\subsubsection{Equilibrium Analysis}
%
%
%
%
%
We are interested in understanding how the introduction of teams may help or hinder cooperation. 
As a first step towards addressing this question, we investigate the impact of teams on the \emph{stage game} of the IPD. 
To provide a clear comparison with the standard IPD, we take an ex-ante approach, where agents are aware of their imminent interaction and the existence of other teams but not the actual team membership of their counterpart.
For further details, refer to Appendix~\ref{sec:appendix_eq_analysis}.

Assume a pair of agents, $i,j$, have been selected to interact at some iteration of the IPD and agent $i$ knows $j$ will be a teammate with probability $\inteam$ and a non-teammate with probability $(1-\inteam)$. 
Let $\sigma_{T_i}=(\sigma_{ji},1-\sigma_{ji})$ represent $j$'s strategy profile when $j \in T_i$, where $\sigma_{ji}$ is the probability for cooperation ($C$).
Likewise, let $\sigma_{T_j}=(\sigma_{jj},1-\sigma_{jj})$ be $j$'s strategy profile when $j \in T_{j}$, any other team.

If agent $i$ decides to cooperate, its expected utility, subject to agent $j$'s strategy, is:
\begin{align}
    \mathbb{E}(C,\sigma_{T}) &= \frac{\inteam (b-c)(\sigma_{ji}+1)}{2}+ (1-\inteam)(\sigma_{jj} b-c).
    \label{eq:team_coop}
\end{align}
If agent $i$ decides to defect, its expected utility, subject to agent $j$'s strategy, is:

\begin{align}
    \mathbb{E}(D,\sigma_{T}) &= \frac{\inteam \sigma_{ji}(b-c)}{2} +(1-\inteam)\sigma_{jj} b.
    \label{eq:team_def}
\end{align}

We determine the conditions under which agent $i$ has incentive to cooperate as when $\mathbb{E}(C,\sigma_{T})\geq  \mathbb{E}(D,\sigma_{T})$.
Substituting Equations~\ref{eq:team_coop} and~\ref{eq:team_def}, this simplifies to:
\begin{align}
    \inteam \geq \frac{2c}{b+c},
    \label{eq:final_requrement}
\end{align}
regardless of $\sigma_{T_i}$ or $\sigma_{T_j}$.
Since $b>c>0$, this constraint is meaningful. 
Satisfying Equation~\ref{eq:final_requrement} shifts the Nash Equilibrium of the stage game from ($D$, $D$) to ($C$, $C$) in expectation.
This means that there exist circumstances where teams can support cooperative incentives, but these circumstances are not universal. 
Our experiments explore this in more detail.

\subsection{Environment 2: Cleanup Markov Game}

Cleanup~\cite{SSDOpenSource} is a temporally and spatially extended Markov game representing a social dilemma.
This domain allows us to examine if the benefits of teams generalize to more complex environments since agents must learn a cooperative policy through movement and decision actions instead of choosing an explicit \emph{cooperation} action like in the IPD.
Active provision is represented in Cleanup by agents choosing actions with no associated environmental reward that are necessary agents to achieve any rewards.

Figure~\ref{fig:cleanup} in Appendix~\ref{sec:cleanup_appendix} shows the Cleanup environment and provides more details about the environment parameters.
At each timestep, agents choose among nine actions: 5 movement (up, down, left, right, or stay), 2 turning (left or right), and a cleaning or punishing beam.
One half of the map contains an aquifer, or river, and the other an apple orchard.
Waste accumulates in the river with some probability at each timestep which must be cleaned by the agents.
Once a cleanliness threshold is reached, apples spawn in the orchard proportional to the overall cleanliness of the river.
Agents receive a reward of +1 for consuming an apple by moving on top of them.
The dilemma exists in agents needing to spend time cleaning the river to spawn new apples and receive no exogenous reward versus just staying in the orchard and enjoying the fruits of another's labor.
Agents have the incentive to stay in the orchard, however if all agents attempt this free-riding policy, no apples grow and no-one gets any reward.
A successful group in Cleanup will balance the temptation to free-ride with the public obligation to clean the river.

\section{Empirical Evaluation}
\label{sec:empirical_eval}

In this section, we present the setup and results of experiments in both environments using MARL agents.
While our teams model does not require it, we assume that for all $T_i,T_j\in\mathcal{T}$, $|T_i|=|T_j|$ (i.e., given a team model, the teams are the same size).
This avoids complications that might arise with agent interactions if teams were significantly different sizes and to be consistent across our domains.
Alternative interaction mechanisms and teams of different sizes are left for future work.
We use the notation $|\mathcal{T}|/|T_i|$ to indicate the total number of teams and the size of each team.
For example, 1/$N$ indicates one team of $N$ agents (fully common-interest) and $N$/1 represents $N$ teams of one agent (fully mixed-motive).
Of course, many scenarios may fall between these two extremes.  
Since fully mixed-motive has agents working as individuals (i.e., no teams), it serves as a benchmark against which we can compare the performance of team structures.

\subsection{IPD Evaluation}

In the IPD, each experiment lasts $1.0 \times 10^6$ episodes where $N=30$ agents learn using Deep $Q$-Learning~\cite{Mnih2015HumanlevelCT}.
An episode is defined by a set of agent interactions where each agent is paired with another agent and plays an instance of the Prisoner's Dilemma.
Agent pairings are assigned using a uniform random distribution over each team so agents are unable to explicitly modify who they interact with, shown as a challenging scenario for cooperation to arise without additional infrastructure~\cite{Anastassacos2020PartnerSF}.
Each experiment is repeated five times.
In Appendix~\ref{sec:ipd_setup}, we prove how this configuration ensures that each agent has the same number of expected interactions to learn from.
Further implementation details are provided in Appendix~\ref{sec:ipd_setup}\footnote{Code: \url{https://github.com/Dtradke/Teams\_IPD}}.

\begin{figure}[t]
    \centering
    \includegraphics[width=\linewidth]{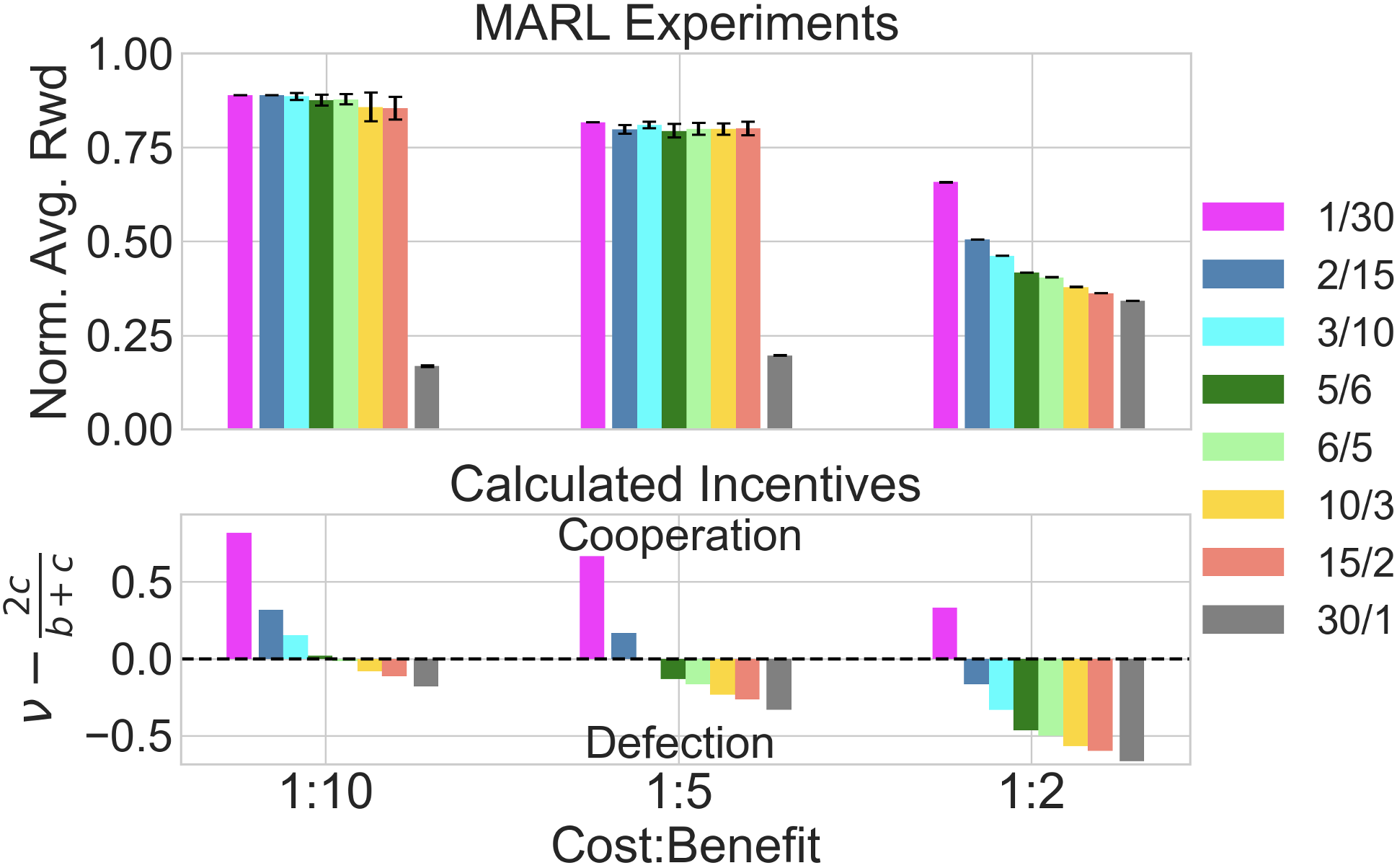}
    \caption{\textbf{IPD:} The top graph shows the normalized average population reward of MARL experiments with three different cost:benefit ratios when $N=30$ with 95\% confidence intervals. The bottom graph shows incentivized actions by Equation~\ref{eq:final_requrement}, where positive (or zero) is cooperation and negative is defection being incentivized. Team structures are labeled $|\mathcal{T}|/|T_i|$ and bookended with fully common interest (1/30) and fully mixed-motive (30/1). When $b \in \{5,10\}$, every team structure besides the individualistic case (30/1) achieves about as much reward as 1/30 without requiring full common interest in the population.}
    \label{fig:teams_adv}
\end{figure}

\subsubsection{Reward}
In our first set of experiments, we explore the degree to which team structures support cooperation.
We fix the cost ($c$) at 1, and let the benefit ($b$) be 2, 5, or 10. 
To capture the behavior of agents after they have converged to a policy, the top graph of Figure \ref{fig:teams_adv} shows the normalized average global reward of the last 25\% of the episodes using MARL agents.
We normalize the average global reward of each experiment in the interval $[0-c,0+b]$ and calculate 95\% confidence intervals to compare different cost and benefit ratios.
To show the corresponding incentives of each experiment, we include the bottom graph which displays the calculated action incentive by a modified Equation~\ref{eq:final_requrement}, $\inteam - \frac{2c}{b+c}$.
Each bar in this graph corresponds with the experiment above so that positive (or zero) bars represent cooperation being incentivized and negative bars represent defection.
Cost and benefit ratios are arranged from highest benefit (left) to lowest benefit (right).

By Equation~\ref{eq:final_requrement}, team structures present agents with the incentive to defect in 13 of 18 scenarios (72\%) in Figure~\ref{fig:teams_adv} (not including 1/30 and 30/1).
Our results show teams always achieve more reward than individual agents (30/1); however, this reward depends on the cost and benefit ratio.
When $b=2$, the experiment results for average population reward a follow trend to the incentives of each scenario in the bottom graph.
Our main finding in Figure \ref{fig:teams_adv} is how, when the benefit increases, MARL agents achieve high average population reward despite the incentive to defect as shown in the bottom graph.
When $b \in \{5, 10\}$, every team structure, other than the individualistic 30/1 scenario, achieves basically the same reward as 1/30 even though there exists mixed-motive environments.
Defection is the incentivized action in seven of 12 (58\%) of these experiments which would produce low global reward if agents actually learned defection.
Instead, we observe agents develop reciprocally pro-social policies that achieve high rewards in every scenario with teams of multiple agents when $b \in \{5, 10\}$.
To analyze how high rewards are achieved in environmental conditions which promote defection, we study agents' behavior over time.

\begin{figure}[t]
    \centering
    \includegraphics[width=\linewidth]{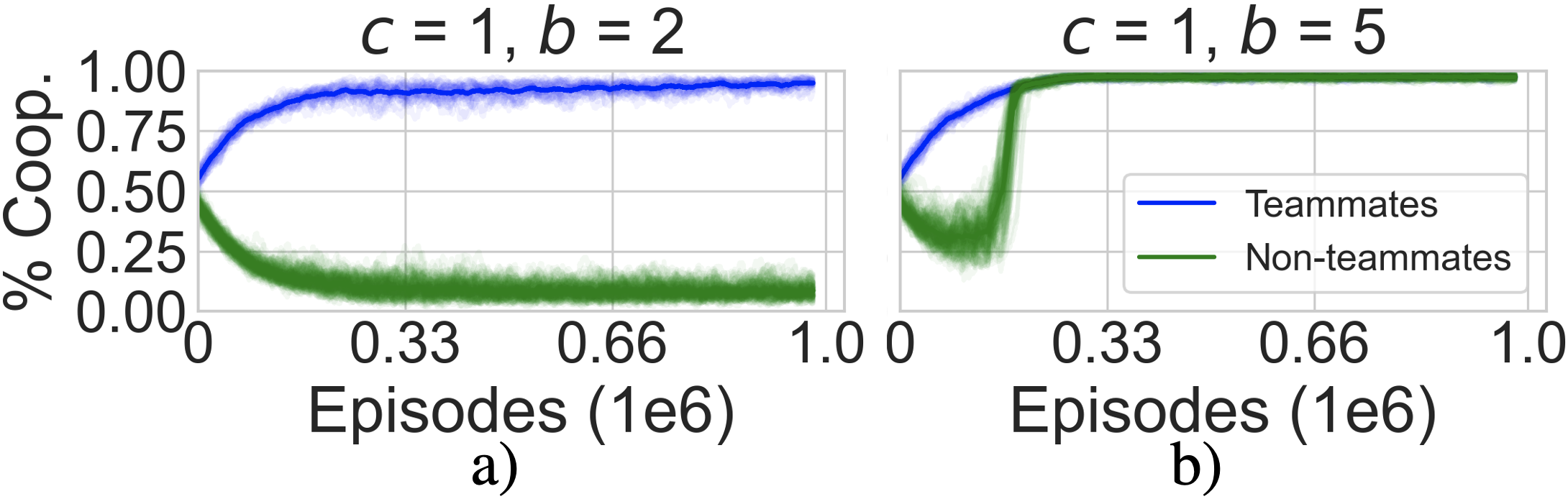}
    \caption{\textbf{IPD:} The 5/6 team composition showing the percent of cooperation towards teammates and non-teammates when $c = 1$ and $b \in \{ 2, 5 \}$. When benefit is greater, agents develop pro-social policies towards non-teammates despite the incentive to defect.}
    \label{fig:teams_adv_overtime}
\end{figure}

\subsubsection{Policy}
In evolutionary biology, fostering cooperation at various \emph{levels} has been found to depend on the size of the cooperative return~\cite{Schnell2021levels}.
Different types of cooperation, or levels of cooperation, have yet to be explicitly explored in MARL.
We identify two levels of cooperation in our IPD environment: with teammates and with non-teammates.
Figure~\ref{fig:teams_adv_overtime} shows the percent of cooperative actions over time with the 5/6 team structure when $b \in \{2, 5\}$.
By Equation \ref{eq:final_requrement}, agents have the incentive to defect in both scenarios.
The $x$-axis shows time and the $y$-axis shows the percent of an agent's actions that are cooperation (2,000 episode averages).

Both graphs in Figure~\ref{fig:teams_adv_overtime} show that agents immediately learn to cooperate with teammates regardless of $b$.
When $b=2$, agents defect on non-teammates; however, when $b=5$, agents learn to cooperate with both teammates and non-teammates.
We observe similar behavior with every other team structure (not including 30/1) when $b \in \{5, 10\}$.
That is, cooperation emerges with teammates and non-teammates despite incentives to defect.
While other work requires strong assumptions of agent behavior to foster cooperation, our results indicate teams allow agents to learn an emergent cooperative convention at multiple levels of a system.

\subsection{Cleanup Evaluation}

In Cleanup, similar to previous work~\cite{Hughes2018InequityAI,McKee2020SocialDA,Jaques2019SocialIA}, we experiment with $N=6$ agents.
Our agents use the Proximal Policy Optimization (PPO)~\cite{PPO2017} RL algorithm for $1.6 \times 10^8$ environmental timesteps (each episode is 1,000 timesteps).
Agent observability is limited to a 15 $\times$ 15 RGB window.
Teammates share the same color and optimize for $TR_i$ calculated at each environmental timestep.
Each experiment is repeated for eight trials.
Further details are in Appendix~\ref{sec:cleanup_appendix}.

\subsubsection{Reward}
\begin{figure}[t]
    \centering
    \includegraphics[width=\linewidth]{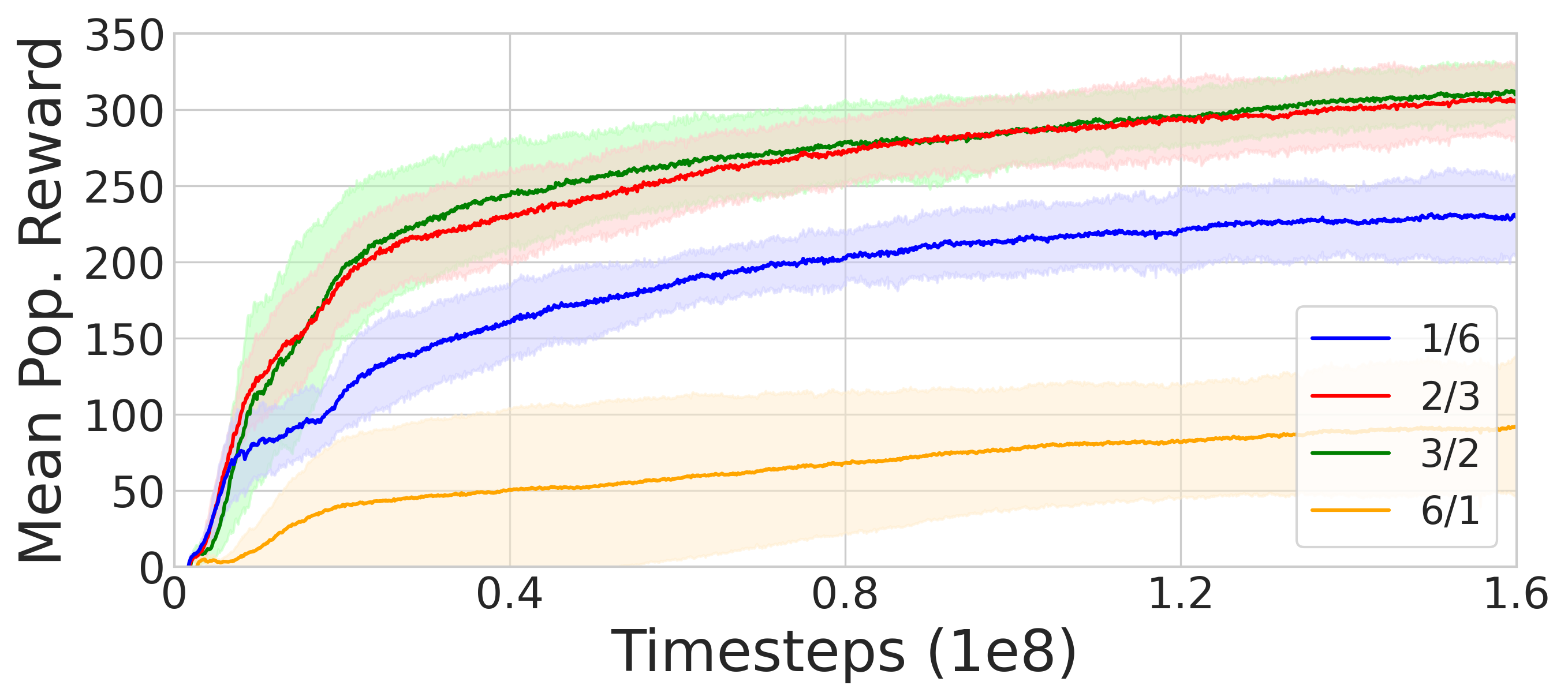}
    \caption{\textbf{Cleanup: }Mean population reward for each team structure with 95\% confidence intervals. 6/1 represents individualistic agents and 1/6 represents when all agents have common interest. Both 2/3 and 3/2 team structures achieve more reward than 1/6 and 6/1.}
    \label{fig:cleanup_rewards}
\end{figure}

Figure~\ref{fig:cleanup_rewards} shows the mean population reward for each scenario in Cleanup with 95\% confidence intervals.
It has been previously shown that the setting that achieves the most population reward is when agents
are altruistic and optimize for the collective rewards of the entire group~\cite{Wang2019EvolvingIM,McKee2020SocialDA}, similar to our 1/6 configuration.
However, teams introduce a new dynamic to the environment and we find the 2/3 and 3/2 team structures both achieve higher reward than 1/6 despite the interests of all agents not being aligned.
As expected, the 6/1 scenario fails to achieve significant reward since agents succumb to the incentive to free ride and few apples grow.
\cite{McKee2020SocialDA} has shown that only evaluating a system for mean reward masks other dynamics such as high levels of reward inequality among agents. 

\subsubsection{Equality}
It is important to consider the process of how teams achieve higher reward and the potential side effects on population equality, such as the reward distributed among agents.
We model population reward equality as the inverse Gini index, similar to past work~\cite{McKee2020SocialDA}, calculated as:

\begin{equation}
    Equality = 1 - \frac{\sum_{i=0}^{N} \sum_{j=0}^{N} |R_i - R_j|}{2|N|^{2} \overline{R_{N}}},
\end{equation}

\begin{figure}[t]
    \centering
    \includegraphics[width=\linewidth]{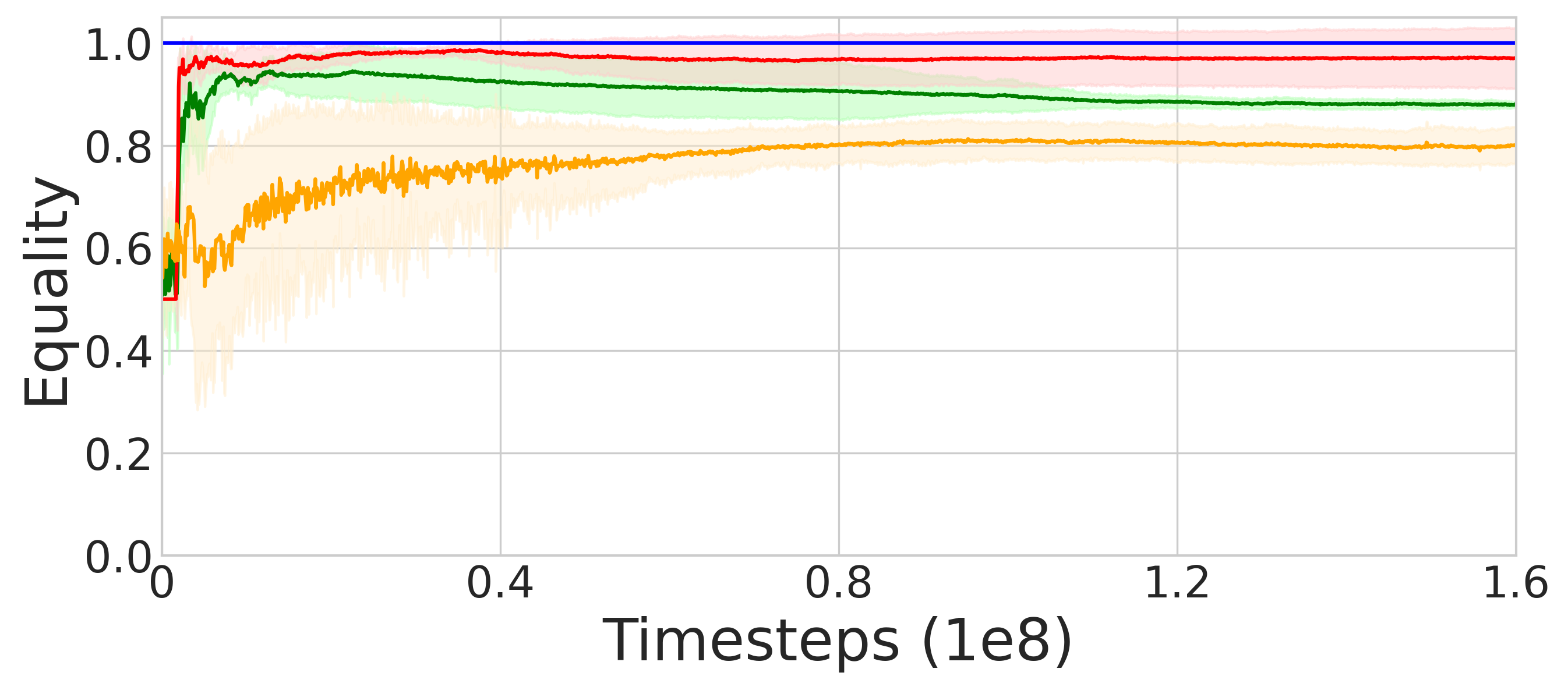}
    \caption{\textbf{Cleanup: }Inverse Gini index (equality) for each team structure with 95\% confidence intervals. Higher values represent more equality. Both 2/3 and 3/2 team structures have high equality despite the interests of all agents not being aligned.}
    \label{fig:cleanup_equality}
\end{figure}

\begin{figure*}[t]
    \centering
    \includegraphics[width=\linewidth]{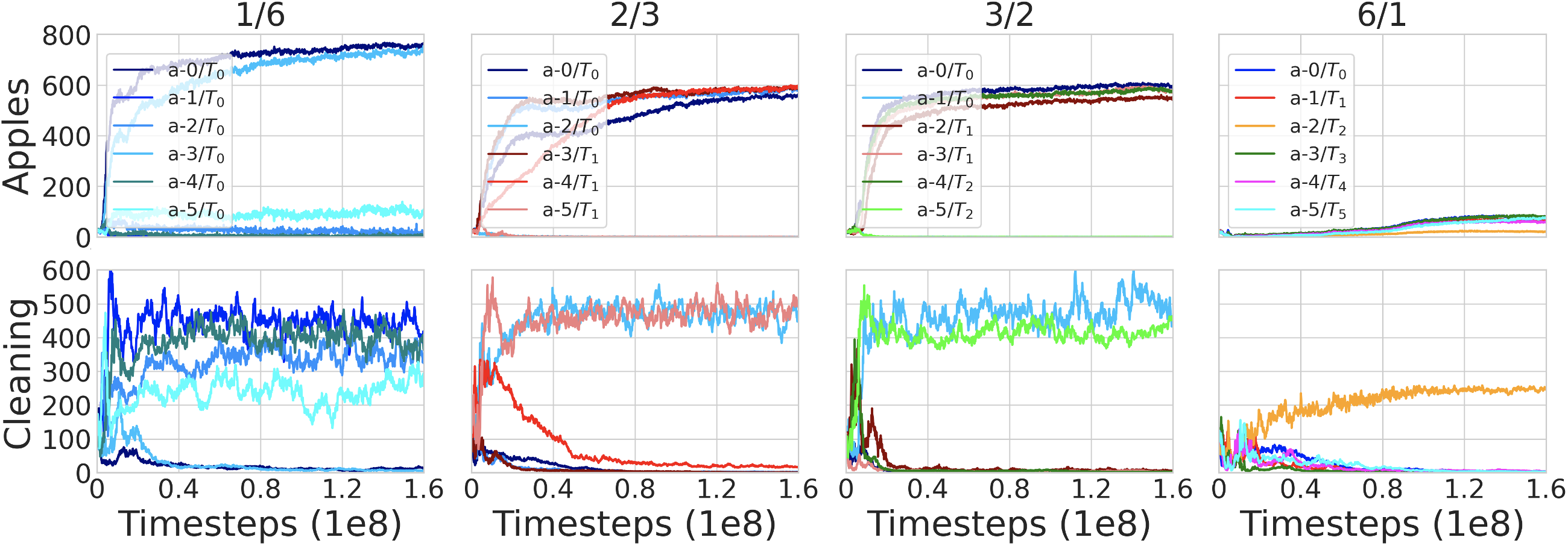}
    \caption{\textbf{Cleanup: }Mean number of apples (top) and cleaning beams (bottom) per-episode for each of 6 agents in different team structures.}
    \label{fig:div_labor}
\end{figure*}

\noindent
where $\overline{R_{N}}$ is the mean population reward.
Figure~\ref{fig:cleanup_equality} shows our results for equality where higher values represent more equality.
The 1/6 scenario is, by definition, always 1 since there is only one team.
Despite earning high reward, both 2/3 and 3/2 team structures also achieve high equality and always have greater equality than 6/1.
Success in Cleanup relies on agents coordinating to form an effective joint policy instead of simply choosing an explicit cooperation action (as in the IPD).
To further understand how team structures achieve the highest rewards while also maintaining high equality, we analyze agents' policies and division of labor among teammates.

\subsubsection{Division of Labor}
While agents consistently learn to divide labor among them, the same numbered agent does not always learn the same behavior across different trials of our experiments which makes aggregating multiple trials difficult.
Therefore, Figure~\ref{fig:div_labor} shows the mean apples picked (top) and cleaning beams (bottom) for each agent in one trial of our evaluation.
The behavior in this trial represents the most common division of labor for each team structure.
Agents on the same team in each plot are presented as different shades of the same color.
The $y$-axis 
shows the number of apples collected or cleaning actions taken and the $x$-axis represents time.
Agents rarely punish, thus we omit it from our analysis.

In the 1/6 configuration (Figure~\ref{fig:div_labor}, left), two agents learn to mostly pick apples while four agents clean the river.
While this represents the most common division of labor with 1/6, we do observe two trials where three agents learn to pick apples and three agents learn to clean the river.
These strategies achieve high mean reward but is not the best division of labor and consistently achieves less reward than the 2/3 and 3/2 team structures.
When analyzing both team structures of 2/3 and 3/2 (Figure~\ref{fig:div_labor}, middle columns), agents tend to divide themselves into four apple pickers and two river cleaners.
This division of labor consistently achieves the highest reward in our evaluation.
The 3/2 team structure tends to learn this division slightly quicker, although on average both configurations eventually achieve basically the same reward as shown in Figure~\ref{fig:cleanup_rewards}.
Independent agents in 6/1 fail to significantly clean the river, therefore few apples grow which leads to low rewards.
In the example shown in Figure~\ref{fig:div_labor}, five agents free-ride on the labor of only one river cleaning agent.

In summary, our results show how agents in team structures learn better task specialization among the population by autonomously learning \emph{roles} within their team.
This allows populations in the 2/3 and 3/2 team structures to keep the river clean while most agents collect the spawning apples.
This causes 2/3 and 3/2 to achieve high population reward and equality across teams even though agents on different teams optimize for their own team's reward.

\section{Discussion and Future Work}
\label{sec:discussion}

Across multiple domains, we have shown that our model of teams has a significant impact on how agents learn to develop pro-social policies and coordinate their behavior.
In the IPD, we show how teams allow agents to immediately identify and cooperate with their teammates, which may be similar to kin selection in human behavior~\cite{Muthukrishna2017Corr}.
Interestingly, we find that RL agents develop a pro-social convention and adapt this cooperative behavior towards non-teammates, even if defection has greater expected value.
This behavior may be comparable with different levels of cooperation in humans, similar to increasing cooperation from only kin selection to notions of direct reciprocity with other groups~\cite{Muthukrishna2017Corr}.


While it was previously thought that optimizing for signals from all agents achieves the highest reward in Cleanup~\cite{Wang2019EvolvingIM,McKee2020SocialDA}, our results show that agents optimizing for only a subset of the population (i.e., a team) achieves higher reward.
Agent specialization in Cleanup is identified in~\cite{McKee2020SocialDA}.
In that work, specialization is viewed as a negative result which causes high labor inequality.
However, the context of teams should change how task specialization is viewed in MARL.
In the literature on Team Forming and Coalition Structure Generation, teams are often explicitly constructed to fill necessary roles~\cite{Andrejczyk2017Concise}.
We view task specialization as the agents autonomously learning these roles with only the feedback of their team's reward.
This reinforces our hypothesis that 
teams can help improve how MARL agents learn to coordinate, and may be of specific interest to the emergent behavior community.


However, certain side effects may occur among teams.
While our 3/2 team structure achieves high reward in Cleanup, there is higher inequality than 2/3.
To achieve the four picker/two cleaner division of labor, one team ($T_{1}$ (red) in Figure~\ref{fig:div_labor}) must free-ride on the labor of the other two teams.
In practice, systems should consider potential side effects if slight inequality is detrimental to its welfare in the long-run, despite short-term stability.
Furthermore, while we explore teams of AI agents, teams may also consist of humans or hybrid populations of both AI and humans.
Exploring alternative team reward functions may lead to interesting results and future research, particularly in the context of hybrid teams.

We see many interesting open questions with multiagent teams.
For example, constructing richer reward structures by individuals optimizing for various types of goals~\cite{radke2022importance}.
Regarding levels of cooperation, exploring teams of unequal size and conditions under which low-level cooperation (i.e., nepotism or bribery) undermines global progress may be of interest.
Our model is constructed to easily allow the adaptation of additional infrastructure among the agents.
Thus, longer term questions include analyzing how features such as communication, negotiation, trust, and sanctions impact our model and introduce new challenges.
We hope that this work will reinvigorate the study of multiagent teams with RL agents to further understand how findings in organizational psychology and AI can complement each other.


\section*{Acknowledgements}
This research is funded by the Natural Sciences and Engineering Research Council of Canada (NSERC), an Ontario Graduate Scholarship, a Cheriton Scholarship, and the University of Waterloo President's Graduate Scholarship.
We also thank Jesse Hoey, Robin Cohen, Alexi Orchard, Sriram Subramanian, Valerie Platsko, and Kanav Mehra for their feedback and useful discussion on earlier drafts of this work.

\bibliographystyle{named}

\appendix
\clearpage
\newpage
\appendix

\section{Equilibrium Analysis of IPD}
\label{sec:appendix_eq_analysis}




\subsection{Expected Utilities}
\label{sec:appendix_utilities}

The expected utility of choosing to cooperate ($C$) or defect ($D$) can be derived using on Table 1, Table 2, $\inteam$, and the strategy profile of $j$, $\sigma_{T}$.
As a first step towards addressing this question, we investigate the impact of teams on the \emph{stage game} of the IPD. 
To provide a clear comparison with the standard IPD, we take an ex-ante approach, where agents are aware of their imminent interaction and the existence of other teams but not the actual team membership of their counterpart. 

Assume a pair of agents, $i,j$, have been selected to interact at some iteration of the IPD and agent $i$ knows $j$ will be a teammate with probability $\inteam$ and a non-teammate with probability $(1-\inteam)$. 
Also assume agent $j$ is playing some strategy summarized by the probability that agent $j$ selects action $C$ conditioned on if they are a teammate or non-teammate.
Let $\sigma_{T_i}=(\sigma_{ji},1-\sigma_{ji})$ where $\sigma_{ji}$ is the probability for action $C$ represent when $j \in T_i$ and $\sigma_{T_j}=(\sigma_{jj},1-\sigma_{jj})$ when $j \in T_{j}$, any other team.
First we show the derivation for $i$'s expected utility for choosing $C$ subject to $j$'s strategy:


\begin{align*}
    &\begin{aligned}
        \mathbb{E}(C,\sigma_{T}) &= \inteam \left[ \sigma_{ji} (b-c) + (1-\sigma_{ji}) \frac{b-c}{2} \right] + \\ &(1 - \inteam) \left[ \sigma_{jj} (b-c) + (1-\sigma_{jj}) (-c) \right] \\ \\
        &= \inteam \left[ \frac{2\sigma_{ji} (b-c)}{2} + \frac{b-c}{2} - \frac{\sigma_{ji} (b-c)}{2} \right] + \\ &(1 - \inteam) \left[ \sigma_{jj} b - \sigma_{jj} c - c + \sigma_{jj} c \right] \\ \\
        &= \inteam \left[ \frac{\sigma_{ji} b - \sigma_{ji} c}{2} + \frac{b-c}{2} \right] + (1 - \inteam) \left[ \sigma_{jj} b - c \right] \\
        &= \inteam \left[ \frac{(b - c) (\sigma_{ji} + 1)}{2} \right] + (1 - \inteam) \left[ \sigma_{jj} b - c \right] \\
        &= \frac{\inteam (b-c)(\sigma_{ji}+1)}{2}+ (1-\inteam)(\sigma_{jj} b-c)
    \end{aligned}
\end{align*}






Now we show the derivation for $i$'s expected utility for choosing $D$ subject to $j$'s strategy:

\begin{align*}
    &\begin{aligned}
        \mathbb{E}(D,\sigma_{T}) &= \inteam \left[ \sigma_{ji} \frac{(b-c)}{2} \right] + (1 - \inteam) \left[ \sigma_{jj} b \right] \\
        &= \frac{\inteam \sigma_{ji}(b-c)}{2} + (1-\inteam)\sigma_{jj} b
    \end{aligned}
\end{align*}



The terms for playing defection with a counterpart who also defects is zero, therefore omitted above.

\subsection{Action Incentives}
\label{sec:coop_incentive}

The moment when agents have the incentive to cooperate given the expected utilities of cooperation and defection is presented in Constraint~\ref{eq:appendix_constraint}.

\begin{equation}
    \mathbb{E}(C,\sigma_{T})\geq  \mathbb{E}(D,\sigma_{T})
    \label{eq:appendix_constraint}
\end{equation}

We calculate this scenario by substituting $\mathbb{E}(C,\sigma_{T})$ and $\mathbb{E}(D,\sigma_{T})$ from above.

\begin{equation*}
\begin{multlined}
    \frac{\inteam (b-c)(\sigma_{ji}+1)}{2}+ (1-\inteam)(\sigma_{jj} b-c) \geq \\ \frac{\inteam \sigma_{ji}(b-c)}{2} +(1-\inteam)\sigma_{jj} b
\end{multlined}
\end{equation*}

\begin{equation*}
    \frac{\inteam (b-c)(\sigma_{ji}+1)}{2} - c + \inteam c \geq \frac{\inteam \sigma_{ji}(b-c)}{2}
\end{equation*}

\begin{equation*}
    \frac{\inteam (b-c)}{2} - c + \inteam c \geq 0
\end{equation*}

\begin{equation*}
    \inteam (b-c) + 2 \inteam c \geq 2c
\end{equation*}

\begin{equation*}
    \inteam b + \inteam c \geq 2c
\end{equation*}

\begin{equation}
    \inteam \geq \frac{2c}{b+c}
    \label{eq:appendix_eq}
\end{equation}

The above derivation simplifies Constraint~\ref{eq:appendix_constraint} to calculate the point at which agents have incentives to cooperate in our environment.
The incentives for each team structure in our IPD environment can be visualized in the bottom graph of Figure~\ref{fig:teams_adv}.

In the regular IPD without teams, agents have no common interest making ($D$, $D$) the unique Nash Equilibrium and ($C$, $C$), ($C$, $D$), and ($D$, $C$) the three Pareto Efficient strategies.
Since teammates share rewards, the degree of common interest is ultimately determined by the amount they interact with their team, $\inteam$.
Therefore if Equation \ref{eq:appendix_eq} is satisfied, the game-theoretical properties of the IPD transform so that ($C$, $C$) is the unique Nash Equilibrium and Pareto Efficient strategy.

\section{Environment Setups}

\subsection{Iterated Prisoner's Dilemma (IPD)}
\label{sec:ipd_setup}

\subsubsection{IPD Payoff Scheme}

Table~\ref{tab:pd_table} shows the Prisoner's Dilemma matrix game used for our IPD experiments and equilibrium analysis.
This parameterization of the IPD considers the cost ($c$) and benefit ($b$) of cooperation where mutual defection yields a reward of 0.
Agents interact with their counterpart and receive the reward from this matrix corresponding with their action and the action of their counterpart.
The team reward $TR_i$ is calculated after all agents in the population interact and is used by agents when learning.

Table~\ref{tab:pd_team} shows how the payoffs of the Prisoner's Dilemma change when two teammates are chosen to interact and fully share rewards.
In this scenario, we observe the unique Nash Equilibrium shift from mutual defection to mutual cooperation.
Agents receive the explicit payoff from Table~\ref{tab:pd_table} from the interaction, though share their rewards with their teammates.
Thus, Table~\ref{tab:pd_team} represents their payoff after $TR_i$ is calculated when interacting with a teammate.

\subsubsection{Matching Algorithm}

At each instance of the IPD, agent $i$ is given a counterpart, $j$ to play the game in Table~\ref{tab:pd_table}.
Agent pairings are assigned using a uniform random distribution from each team, meaning the probability of a counterpart being chosen from $T_i$ is the same as any other team $T_j$.
For example, if $|\mathcal{T}| = 5$ an agent has a 20\% chance of being paired with a teammate.
Each episode, we construct $N$ pairings by matching each agent $i\in N$ to a partner $j\in N\setminus\{i\}$, with the constraint that the probability of $j$ being on any team is equally likely.
Each agent observes the team their counterpart belongs to through a numerical signal $s_i \in S$, but not their actual individual identity.

\subsubsection{Team Size and the Number of Interactions}

In each episode, agents are given a counterpart and could also be chosen to be the counterpart of another agent for a total of $N$ pairings per-episode.
Since agents learn only through their own direct interactions, we must ensure that the particular matching process we use does not bias the results. 
In particular, we need to be confidant that the underlying team structure in which agents are embedded in no way influences the agent training through under- or over-sampling or providing disproportionate opportunities to be matched and play an iteration of the IPD. 

\begin{prop}
If $|T_i| = |T_j|$ $\forall i, j \in N$ and agents are randomly paired from any team with uniform probability, each agent will have the same expected number of IPD interactions for any value of $|T|$ or $N$.
\label{thm:same_plays}
\end{prop}

\begin{proof}
Let a population of $N$ agents be split up into $|\mathcal{T}|$ teams of size $n$, so that $N = |\mathcal{T}|n$.
Since agents are paired with an agent from any team with equal probability, $Pr(IN) = 1 - \frac{1}{|\mathcal{T}|(n-1)}$ and $Pr(OUT) = 1 - \frac{1}{|\mathcal{T}|n}$ represents the probability of \textbf{not} being matched with a teammate or non-teammate respectively.
These are different since an agent is unable to be paired with themselves, leaving $n-1$ agents to possibly be paired with from their own team.
The probability of agent $i$ not being chosen as the matching agent is defined as:

\begin{equation*}
    Pr(\overline{i})_{|\mathcal{T}|n} = Pr(IN)^{n-1} + Pr(OUT)^{n(|\mathcal{T}|-1)}.
\end{equation*}

Suppose $m$ agents are added to each team so that $N' = |\mathcal{T}|n + |\mathcal{T}|m$ and $n := n + m$.
In this new setting, the probability of $i$ not being chosen in a population of $|\mathcal{T}|(n+m)$ agents becomes:

\begin{equation*}
\begin{multlined}
    Pr(\overline{i})_{|\mathcal{T}|(n+m)} =  \\Pr(IN)^{(n-1)+m} + Pr(OUT)^{n(|\mathcal{T}|-1)+(|\mathcal{T}|m-m)}.
\end{multlined}
\end{equation*}

We can derive that $Pr(\overline{i})_{|\mathcal{T}|(n+m)} - Pr(\overline{i})_{|\mathcal{T}|n} = (|\mathcal{T}|m-m)+m$, which simplifies to $|\mathcal{T}|m$.
Note that $N' - N = |\mathcal{T}|m$ also. 
While the probability of not being chosen increases by $|\mathcal{T}|m$, the total interactions in each episode also increases by $|\mathcal{T}|m$.
Thus, agents have the same number of expected interactions.
\end{proof}

\begin{table}[t]
\begin{center}
 \begin{tabular}{|c||c|c|} 
 \hline
  & Cooperate & Defect \\ [0.5ex] 
 \hline\hline
Cooperate & $b-c$, $b-c$ & $-c$, $b$ \\\hline
Defect & $b$,  $-c$ & 0, 0 \\
 \hline
\end{tabular}
\caption{An example of the Prisoner's Dilemma with the costs (c) and benefits (b) of cooperating ($b>c>0$).}
\label{tab:pd_table}
\end{center}
\end{table}

\begin{table}[t]
\begin{center}
 \begin{tabular}{|c||c|c|} 
 \hline
  & Cooperate & Defect \\ [0.5ex] 
 \hline\hline
Cooperate & $b-c$, $b-c$ & $\frac{b-c}{2}$, $\frac{b-c}{2}$ \\\hline
Defect & $\frac{b-c}{2}$,  $\frac{b-c}{2}$ & 0, 0 \\
 \hline
\end{tabular}
\caption{An example of the Prisonner's Dilemma when agents are teammates. $(C,C)$ is the unique Nash Equilibrium.}
\label{tab:pd_team}
\end{center}
\end{table}


Proposition \ref{thm:same_plays} says if each team in $\mathcal{T}$ is the same size and counterparts are randomly chosen from teams with uniform probability, each agent will have the same expected number of interactions to train their policies.
Intuitively, while the probability of being selected as a counterpart decreases as $|T|$ or $N$ increases, there are more opportunities to be chosen.
Note that this result could also be obtained with teams of different sizes so long as the pairing probability is distributed appropriately.
This helps ensure our empirical results are attributed to the dynamics of multiagent teams instead of inherent bias favoring agents with more experience.
We denote the expected number of interactions as $I$ for our analysis in following analyzes.

\subsubsection{IPD Reinforcement Learning Algorithm}
\label{subsec:MARL_exp}

For each round of the IPD, agent $i$ is paired with another agent $j$ chosen randomly from the population.
The two agents play one iteration of the game shown in Table \ref{tab:pd_table}.
Each agent observes the team their counterpart belongs to instead their actual identity. In particular,  for a pair of agents, $i$ and $j$,  their states $s_i$ and $s_j$ are defined as   $s_i = T_j$ and $s_j = T_i$.
Given each $s$, the two agents simultaneously choose an action $a$ which is whether to cooperate or defect.
They do not observe the action their counterpart takes, but instead receive rewards $TR_i$ and $TR_j$ based on their own interaction and those of their teammates.
Each $i$ (and also $j$ with their information) stores the tuple $\langle s_i, a_i, TR_i \rangle$ in their replay buffer to train their policy after each episode using Deep $Q$-Learning~\cite{Mnih2015HumanlevelCT} by sampling a random batch of 32 interactions.
Each agent's internal neural network consists of an input layer of size $|\mathcal{T}|$, two hidden layers of 200 nodes each with hypobolic tangent activation functions, and a two-action output layer with a linear activation function.
Our agents use a learning rate of $1 \times 10^{-4}$ and discount factor of $0.99$ with $\epsilon$-exploration.

\subsection{Cleanup Markov Game}
\label{sec:cleanup_appendix}

\begin{figure}[t]
    \centering
    \includegraphics[width=\linewidth]{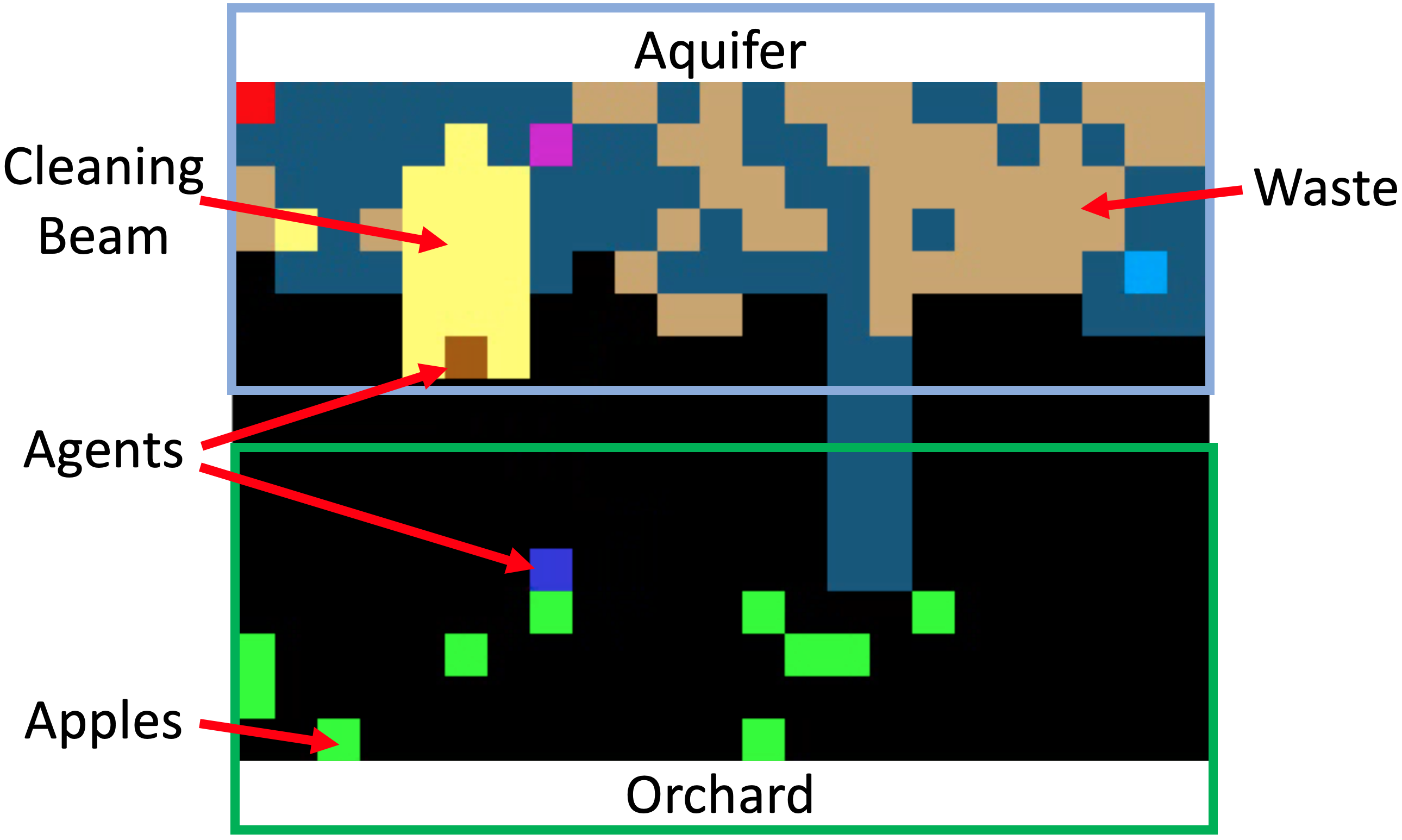}
    \caption{Cleanup environment with 6 agents and no teams.}
    \label{fig:cleanup}
\end{figure}

\subsubsection{Cleanup Reinforcement Learning Algorithm}
\label{sec:cleanup_rl_alg}

The typical environmental setup for Cleanup implemented in past work uses five agents.
However, this would only create 1/5 and 5/1 when considering multiple possible teams of equal size.
Therefore, we instantiate six agents to create more team structures.
We deploy the default Proximal Policy Optimization (PPO)~\cite{PPO2017} learning algorithm architecture in the Cleanup repository~\cite{SSDOpenSource}.
PPO is a policy gradient algorithm which constrains the space of policy updates to avoid large policy updates for smoother training and has been previously shown as a good algorithm for agents to learn in Cleanup.
Agents are only able to observe the a $15 \times 15$ box centered at their location and update their policies using the environment's default batch size of at least 16,000 timesteps and a maximum of 100,000 timesteps. 
Each episode executes for 1,000 timesteps and we run experiments for $1.6 \times 10^8$ timesteps.
The learning rate decreases linearly from $1.2 \times 10^{-3}$ to $1.2 \times 10^{-5}$ over the first $2 \times 10^7$ timesteps and remains static at $1.2 \times 10^{-5}$ afterwards.

\subsubsection{Default Environment Parameters}

An example of Cleanup is shown in Figure~\ref{fig:cleanup}.
The social dilemma dynamics of Cleanup rely on the existence of waste and apples.
The functions which govern the creation of these features can be easily modified; however, we evaluate our model of teams primarily using the default parameters of the environment which include a waste regeneration rate, waste regeneration threshold, and apple generation rate.
Once less than 40\% of the river (aquifer) grid-cells contain waste, waste regenerates at each clean cell with a probability of 50\% at each timestep.
Below this waste regeneration, apples spawn at each location in the orchard with a linear probability ranging from 0\% when waste is 40\% of the river up to 5\% when waste makes up 0\% of the river.

We explored a 20\% waste threshold and 2.5\%, 10\%, and 20\% maximum apple regeneration probability and found no significant alterations in the results.
The best joint strategy among the six agents remained four pickers and two cleaners and both 2/3 and 3/2 team structures performed best.




\end{document}